\documentclass[runningheads]{llncs}
\usepackage[T1]{fontenc}
\usepackage{graphicx}
\usepackage{marvosym} %这个包用来给通信作者生成一个“Letter”标志
\usepackage{amssymb}
\usepackage{amsmath}
\usepackage{multirow}
\usepackage{booktabs}
\usepackage[linesnumbered,ruled,vlined]{algorithm2e}
\begin{document}
\title{Learn to Unlearn: Meta-Learning-Based Knowledge Graph Embedding Unlearning}
\titlerunning{Learn to Unlearn: Meta-Learning-Based KGE Unlearning}
\author{Naixing Xu\inst{1}\orcidID{0009-0005-2094-330X}\and
Qian Li \inst{2} \and
Xu Wang \inst{1}\orcidID{0009-0000-1656-1572} \and
Bingchen Liu \inst{1}\orcidID{0000-0001-5041-8593}
Xin Li\inst{1}\textsuperscript{(\Letter)}\orcidID{0000-0003-2411-5374} }
% 在对应的通讯作者这里加 \inst{(}\Envelope\inst{)}

\authorrunning{N. Xu et al.}
\institute{School of Software, Shandong University, Jinan, China \\
\email{naixingxu@mail.sdu.edu.com}
\and
Joint SDU-NTU Centre for Artificial Intelligence Research (C-FAIR), Shandong University, Jinan, China
}

\maketitle              % typeset the header of the contribution
\begin{abstract}
Knowledge graph (KG) embedding methods map entities and relations from knowledge graphs to continuous vector spaces, simplifying their representations and enhancing performance across various tasks (e.g., link prediction, question answering). As concerns about personal privacy rise, machine unlearning (MU), an emerging AI technology that enables models to eliminate the influence of specific data, has garnered increasing attention from the academic community. Existing works typically achieves machine unlearning through data obfuscation and adjustments to the model's training loss. Furthermore, existing approaches lack generalization ability across different unlearning tasks. In this paper, we propose a \underline{Meta}-Learning-Based Knowledge Graph \underline{E}mbedding \underline{U}nlearning framework (MetaEU), specifically designed for KG embedding unlearning. By leveraging meta-learning, we generate embeddings that require unlearning. This process reduces the impact of specific knowledge on the graph while maintaining the model's performance on the remaining data. A thorough experimental study on benchmark datasets shows that MetaEU demonstrates promising performance in the knowledge graph embedding unlearning task.

\end{abstract}
\section{Introduction}
In recent years, Knowledge graphs (KGs) have emerged as essential structures for representing relational data in various domains, from natural language processing to recommendation systems~\cite{yingTwoStageKnowledgeGraph2024}. Knowledge graph embedding (KGE) techniques, play a crucial role in transforming complex graph structures into low-dimensional vector spaces. These techniques are widely applied in common KG tasks such as knowledge graph completion~\cite{yingTwoStageKnowledgeGraph2024,zhaoAttentionBasedAggregationGraph2020,yuanKnowledgeGraphEmbedding2019}, entity alignment~\cite{xieImprovingKnowledgeGraph2023}, link prediction, etc. Meta-learning first appears in the field of educational psychology. It is defined as the comprehension and adaptation to the process of learning, rather than simply the accumulation of subject knowledge~\cite{tianMetalearningApproachesLearninglearn2022}. In the past few years, researchers have been attempting to apply meta-learning in various fields of artificial intelligence to enhance the generalization ability of existing models and improve their few-shot learning capabilities~\cite{tianMetalearningApproachesLearninglearn2022,chenMetaReinforcementLearningAlgorithm2024}.

However, as KG embeddings are applied to real-world applications, the demand for updating or removing outdated and incorrect content from knowledge graphs is increasing. Additionally, the enactment of privacy protection laws, such as the General Data Protection Regulation (GDPR) and the California Consumer Privacy Act (CCPA), grants users the right to delete their personal data. This also means that models need to be able to eliminate the influence of specific data, a task known as Machine Unlearning. While it is relatively straightforward to remove the data that needs to be unlearned from the training set, the impact of that data remaining in the model is challenging to eliminate. Furthermore, retraining the model from scratch on a new dataset is very time-consuming and resource-intensive. Additionally, due to the interrelationships among entities, ensuring that a model that has undergone unlearning performs well on the remaining data presents a significant challenge.

\begin{figure}
  % \vspace{-0.5cm} % 减少图形和文字之间的间隔
  \centering
  \includegraphics[width=0.8\textwidth]{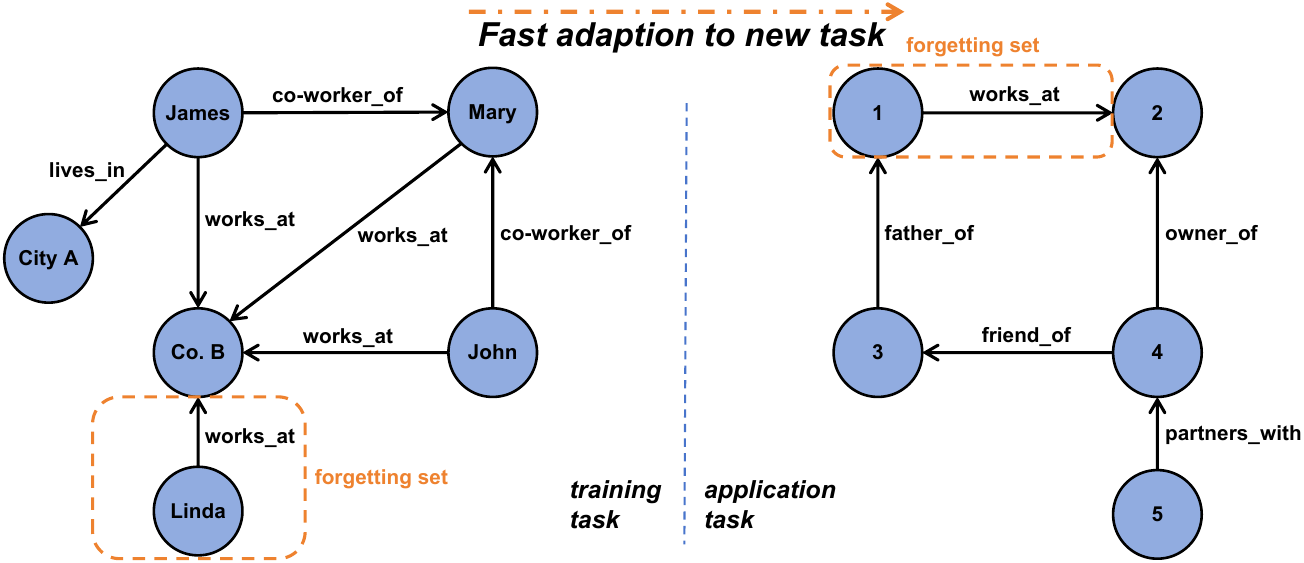}
  \caption{An example of meta-learning-based knowledge graph embedding unlearning.}
  \label{fig1}
  % \vspace{-0.5cm} % 减少图形和文字之间的间隔
\end{figure}

To address this, a meta-learning-based framework for knowledge graph embedding unlearning is proposed. Its goal is to quickly remove the influence of specific data from the model without compromising overall performance, while also generalizing to other unlearning scenarios. As shown in Fig.~\ref{fig1}, with the help of meta-learning-based KG embedding unlearning, social platforms can better protect users' privacy. For example, \textit{Linda} works at \textit{Company B} but wishes to delete this information for privacy reasons. In another scenario, \textit{User 1} also wants to remove similar information. With meta-learning, the model can quickly complete machine unlearning tasks across different scenarios. One existing work~\cite{zhuHeterogeneousFederatedKnowledge2023}, based on cognitive neuroscience theory, achieves specific knowledge unlearning in federated KG embedding through retroactive interference and passive decay. Another approach~\cite{liuFederatedKnowledgeGraph2024} uses a diffusion model to generate noisy embeddings to replace the embeddings that need to be unlearned, thereby achieving specific knowledge unlearning. However, the models proposed in these works can only perform machine unlearning in specific scenarios, and the data generation process in diffusion models is highly complex and slow. These limitations restrict the generalization ability of these works and hinder their ability to perform machine unlearning tasks across different scenarios.

To address the above issues, we propose a novel framework, a \underline{Meta}-Learning-Based Knowledge Graph \underline{E}mbedding \underline{U}nlearning framework (MetaEU). Our framework generates high-quality embeddings to replace the embeddings of knowledge that needs to be unlearned. Consequently, it effectively mitigates the impact of the targeted knowledge while preserving the overall performance of the KG embedding model, enabling machine unlearning in knowledge graphs. 

To achieve \textit{Learn to Unlearn}, MetaEU is designed to learn the meta-attributes of the knowledge (entities and relations) to be forgotten. Specifically, these meta-attributes include relation patterns independent of the entity and the information of the entity's neighborhood structure. We utilize the Relation-Aware Entity Embedding Generator (RAEEG) and the Neighbor-Enhanced Embedding Modulator (NEEM) to capture these meta-attributes. Following the meta-training regime, during the training process of MetaEU, we sample a set of tasks from the training set, each containing a support set and a query set. The entities within these tasks are considered unseen, simulating the machine unlearning tasks across different scenarios. In a single task, the entity embeddings generated by RAEEG and NEEM based on the support set are evaluated on the query set. Through meta-training, the machine unlearning capability of MetaEU can generalize to other scenarios, allowing it to efficiently eliminate the impact of specific knowledge across different knowledge graphs, without the need for complex adaptations to individual scenarios.

Generally, our primary contributions are as follows:
\begin{itemize}
    \item To our best knowledge, this is the first attempt to apply meta-learning to KG embedding unlearning.
    \item We propose MetaEU, a novel framework based on meta-learning for KG embedding unlearning. This framework integrates the Relation-Aware Entity Embedding Generator (RAEEG) and the Neighbor-Enhanced Embedding Modulator (NEEM) to generate high-quality obfuscated data, which dilutes the representations of the knowledge to be unlearned.
    \item We conduct extensive experiments on benchmark datasets, demonstrating the effectiveness of our proposed framework.
\end{itemize}

The remaining parts of this paper are structured as follows: In Section 2, we introduce the related work. In Section 3, we present forumulation of the problem. In Seciton 4, we specify our proposed framework. In Section 5, we conduct experiments and discuss the experimental results. At last, we give the  conclusion of this paper in Section 6. 

\section{Related Work}
In this section, we discuss related work in three key areas: knowledge graph embedding models, meta-learning, and machine unlearning. 

\textbf{Knowledge graph embedding} Knowledge graph embedding (KGE) models are designed to represent entities and relations of a knowledge graph in a continuous vector space to facilitate downstream machine learning tasks.The foundational work by Bordes et al. ~\cite{bordesTranslatingEmbeddingsModeling2013} introduced TransE, a translational distance model that became the basis for many subsequent approaches. Building on this, other models such as DistMult ~\cite{yangEmbeddingEntitiesRelations2014} and ComplEx ~\cite{trouillonComplexEmbeddingsSimple2016} extended the capability of KGE by modeling more complex relational patterns, such as symmetric and antisymmetric relations. Furthermore, Sun et al. ~\cite{sunRotatEKnowledgeGraph2018} proposed RotatE, introducing a rotation-based approach to model complex relations effectively. To address the challenge of modifying KG embeddings after deployment, the latest work in 2024 by Cheng et al. ~\cite{chengEditingLanguageModelBased2024} proposed KGEditor, a framework designed to adjust knowledge graph embeddings.

\textbf{Meta-learning} Meta-learning, or "learning to learn," seaks to improve the generalization ability of models by utilizing experiences from multiple related tasks. The work of Finn et al. ~\cite{finnModelAgnosticMetaLearningFast2017} introduced Model-Agnostic Meta-Learning (MAML), which quickly adapts models to new tasks with limited data by learning an initial set of parameters suitable for fine-tuning. Li et al. ~\cite{liMetaSGDLearningLearn2017} extended this approach by focusing on gradient-based optimization improvements to further enhance convergence speed. The combination of meta-learning with other methods in machine learning, has received increasing attention. Zintgraf et al. ~\cite{zintgrafFastContextAdaptation2019} proposed a meta-learning method that improved adaptability in reinforcement learning settings, while Qiao et al. ~\cite{qiaoDualchannelSemisupervisedLearning2024} applied meta-learning to graph data, addressing the scarcity of supervised information by integrating labeled and unlabeled data on the graph.

\textbf{Machine unlearning} Machine unlearning is an emerging field aimed at efficiently removing specific learned information from machine learning models to address data privacy concerns. Cao and Yang ~\cite{caoMakingSystemsForget2015} laid the groundwork for this concept by proposing a data deletion method that recalculates model parameters in a computationally feasible way. Ginart et al. ~\cite{ginartMakingAIForget2019} developed an efficient unlearning approach specifically for k-means clustering, allowing selective data forgetting. More recently, Kim and Woo ~\cite{kimEfficientTwostageModel2022b} introduced a two-stage model retraining method that utilizes knowledge distillation to enable the rapid removal of specific data without affecting the performance of the deep learning model.  In a 2024 study, Yao et al. ~\cite{yaoMachineUnlearningPretrained2024} proposed a machine unlearning framework in the context of large language models (LLMs), demonstrating that machine unlearning is a viable solution to address the "right to be forgotten" issue within LLMs.

These three areas—knowledge graph embeddings, meta-learning, and machine unlearning—collectively establish the necessary background for our work. Despite the significant advancements achieved in each of these fields, research on the application of machine unlearning within the context of knowledge graphs remains scarce. Our proposed approach seeks to bridge concepts from these domains to advance knowledge representation and its ethical application in machine learning.

\section{Problem Definition}
A KG $\mathcal{G}$ can be considered as a directed graph composed of a series of triples. Let $\mathcal{E}$ is the set of entities, $\mathcal{R}$ is the set of relations, KG can be defined as $\mathcal{G} = (\mathcal{E}, \mathcal{R}, \mathcal{T})$, where $\mathcal{T} = \{(h, r, t)\} \subseteq \mathcal{E} \times \mathcal{R} \times \mathcal{E}$. KGE methods continually train and adjust the entity embedding matrix $\mathbf{E}\in\mathbb{R}^{|\mathcal{E}|\times d}$ and the relation  embedding matrix $\mathbf{R}\in\mathbb{R}^{|\mathcal{R}|\times d}$ through their own scoring function $s(\cdot)$, until both $\mathcal{E}$ and $\mathcal{R}$ have suitable embeddings in the context of the knowledge graph $\mathcal{G}$, where $|\mathcal{E}|$ and $|\mathcal{R}|$ denote the numbers of entities and relations, respectively, and $d$ is the dimension of the embeddings. In other words, the score $s(h,r,t)$ for a true triple $(h,r,s)\in\mathcal{T}$ is higher than the score $s(h',r',t')$ for a negative triple $(h^{\prime},r^{\prime},t^{\prime})\notin \mathcal{T}$. 

To address the problem of achieving machine unlearning in KGE, we introduce a scenario where the knowledge graph is partitioned into two subsets: the forgetting set $\mathcal{T}_f$ and the remaining set $\mathcal{T}_r$. $\mathcal{T}_f$ represents triples that need to be unlearned, while $\mathcal{T}_r$ consists of the rest of the knowledge graph. The objective of machine unlearning in KGE is to rapidly remove the influence of triples in $\mathcal{T}_f$ while maintaining the model's performance on $\mathcal{T}_r$. 

The goal of the meta-learning-based KGE unlearing framework is to leverage the ability of meta-learning to generalize across different unlearning tasks. For instance, in an e-commerce platform, many users may be sensitive about their medicine purchase records and request the platform to delete this data. A meta-learning-based KGE unlearning framework can effectively remove the impact of these records for each user who makes such a request. Specifically, we formulate the unlearning problem as a bi-level optimization process where:
\begin{enumerate}
    \item During meta-training, the model trains on various unlearning tasks, each involving a different subset of triples, to acquire meta-knowledge $\mathcal{M}$ across diverse unlearning scenarios.
    \item During meta-testing, the model leverages the learned meta-knowledge to efficiently adapt embeddings for $\mathcal{T}_f$, producing $E^{\prime}$ that retains suitable properties on the $\mathcal{T}_r$ while minimizing the influence of $\mathcal{T}_f$.
\end{enumerate}

Formally, given a knowledge graph $\mathcal{G}=(\mathcal{E},\mathcal{R},\mathcal{T})$ with $\mathcal{T}=\mathcal{T}_f\cup\mathcal{T}_r$, we want to find a optimal KGE unlearning function:
\begin{equation}
    \mathcal{F}_u(\mathbf{E},\mathcal{T}_f,\theta) \rightarrow \mathbf{E}^{\prime},
\end{equation}
where $\mathcal{F}_u(\cdot)$ is the KGE unlearning function, and $\theta$ is the parameter set. Moreover, the output of the function, $\mathbf{E}^{\prime}$, should satisfy the following properties:
\begin{enumerate}
    \item The performance of embedding matrix $\mathbf{E}^{\prime}$ on the $\mathcal{T}_r$ should be comparable to the performance of $\mathbf{E}$ trained using the KGE method, while exceeding the performance of $\mathbf{E}_r$ obtained by re-training using the KGE method on the $\mathcal{T}_r$.
    \item Meanwhile, the performance of $\mathbf{E}^{\prime}$ on the $\mathcal{T}_f$ should be inferior to both $\mathbf{E}$ and $\mathbf{E}_r$, demonstrating that $\mathbf{E}^{\prime}$ has better eliminated the influence of the data that needs to be forgotten.
\end{enumerate}

\section{Methodology}
\begin{figure}
  % \vspace{-0.5cm} % 减少图形和文字之间的间隔
  \centering
  \includegraphics[width=0.9\textwidth]{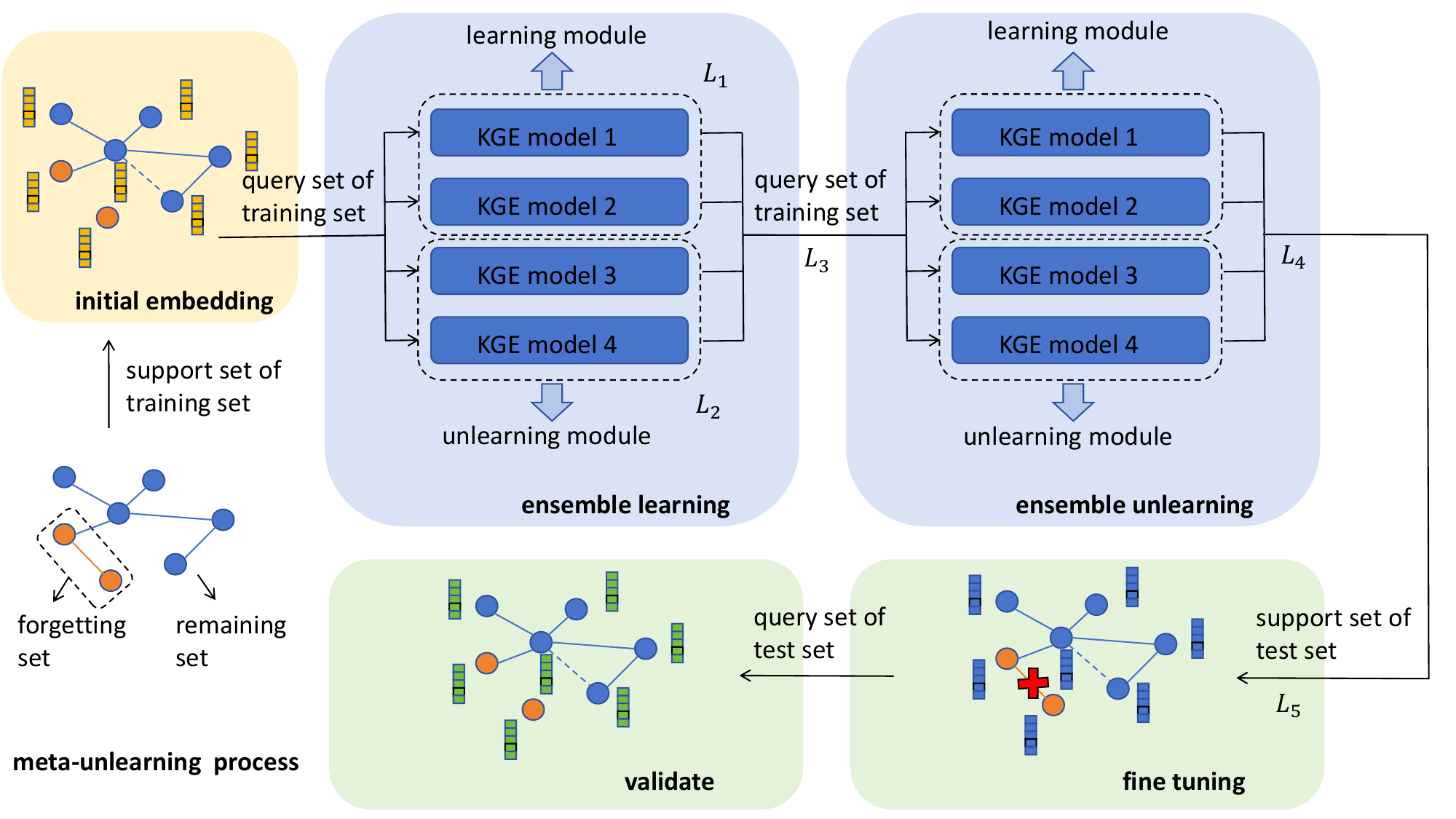}
  \caption{MetaEU innovatively incorporates the processes of ensemble learning and ensemble unlearning within the meta-learning framework.}
  \label{fig2}
  % \vspace{-0.5cm} % 减少图形和文字之间的间隔
\end{figure}

We now present the proposed MetaEU model, which aims to efficiently unlearn specific data from a knowledge graph embedding model while maintaining performance on the remaining data. As shown in Fig.~\ref{fig2}, our framework leverages an ensemble learning strategy that combines multiple base learners, each consisting of two core modules: Relation-Aware Entity Embedding Generator (RAEEG) and Neighbor-Enhanced Embedding Modulator (NEEM). Next, we will provide a detailed explanation of each component of the model.
\subsection{The Framework of Meta-Learning}\label{sec:framework of meta-learning}
The objective of traditional machine learning is to train a model that performs well on a fixed, predefined task. This approach focuses on optimizing a single model for a specific dataset or problem domain. In contrast, meta-learning aims to train a meta-model that can generalize across various tasks. The purpose of MetaEU is to achieve efficient KGE unlearning across various scenarios, which requires the model to generate high-quality obfuscated data even for unseen entities. To implement meta-learning in MetaEU, we need to modify the existing KGE framework to align with the meta-training regime.
\subsubsection{The construction of the Dataset}
In traditional KGE methods, the model typically relies on a dataset $\mathcal{T}$ composed of triples. This dataset $\mathcal{T}$ is usually divided into two parts: a training set $\mathcal{T}_{train}$ and a test set $\mathcal{T}_{test}$ (for simplicity, the validation set is ignored in the discussion). To enhance the model's performance, $\mathcal{T}_{train}$ not only includes the original positive triples $\{h,r,t\}$ but also incorporates negative triples $\{h^{\prime},r^{\prime},t^{\prime}\}$. The model is trained using the following loss function (taking the TransE ~\cite{bordesTranslatingEmbeddingsModeling2013} loss function as an example):
\begin{equation}
    {\mathcal L}=\sum_{(h,r,t)\in \mathcal{T}_{train}}\left[\max(0,\gamma+s(h^{\prime},r^{\prime},t^{\prime})-s(h,r,t))\right],
    \label{KGE loss}
\end{equation}
where $s(h,r,t)=-\|\mathbb{E}_h+\mathbb{R}_r-\mathbb{E}_t\|$ represents the scoring function, and $\gamma$ denotes the margin hyperparameter.

According to the concept of meta-learning, in MetaEU, we extract $k$ subgraphs from the KG $\mathcal{G}$ and treat the entities within these subgraphs as unseen to simulate unknown task scenarios:
\begin{equation}
    \mathcal{G}_i=(\mathcal{E}_i,\mathcal{R}_i,\mathcal{T}_i),\quad\forall i\in\{1,2,\ldots,k\},
\end{equation}
where $\mathcal{E}_i$ represents the entity set of the subgraph $\mathcal{G}_i$, $\mathcal{R}_i$ represents the relation set of the subgraph $\mathcal{G}_i$, and $\mathcal{T}_i \subseteq \mathcal{E}_i \times \mathcal{R}_i \times \mathcal{E}_i$ denotes the triple set of the subgraph $\mathcal{G}_i$. Additionally, a portion of the triples in these subgraphs is used as the support set, allowing the model to generate appropriate embeddings, while the remaining triples form the query set, which is used to evaluate the quality of the generated embeddings and compute the training loss function:
\begin{equation}
    \mathcal{T}_{i}=\mathcal{T}_{i}^{support}\cup\mathcal{T}_{i}^{query},\quad\mathcal{T}_{i}^{support}\cap\mathcal{T}_{i}^{query}=\emptyset.
    \label{eq:support set query set}
\end{equation}

To adapt MetaEU to different task scenarios, we treat the entities within the task subgraph $\mathcal{G}_i$ as unseen entities. By training on these tasks, MetaEU naturally exhibits strong generalization capabilities.

\subsubsection{The Procedure of Meta-Learning}
After obtaining $k$ task subgraphs, MetaEU performs learning on these subgraphs to enable knowledge transfer to new task scenarios. According to meta-learning theory, the model acquires meta-knowledge during this process. Meta-knowledge can be regarded as a form of high-level structure or rules shared across tasks, which we will elaborate on in Section 4.2. For each task $T_i$ and its corresponding task subgraph $\mathcal{G}_i$, MetaEU generates embeddings using the support set $\mathcal{T}_i^{support}$ and evaluates them on the query set $\mathcal{T}_i^{query}$. The overall objective function of meta-learning can be expressed as: 
\begin{equation}
    \min_{\phi}\mathbb{E}_{T\sim p(T)}\left[\mathcal{L}_{\mathcal{T}_{i}^{support}}(f_{{\phi}^{\prime}}(x_{j}),y_{j})\right],
\end{equation}
where $p(T)$ represents the task distribution, and each task $T$ is a specific task sampled from this distribution; the function $f_{\phi}(\cdot)$ is an abstract representation of the meta-knowledge component in MetaEU, which is expected to perform well on the query set $\mathcal{T}_i^{query}$ after observing the support set $\mathcal{T}_i^{support}$; $\mathcal{L}_{\mathcal{T}_{i}^{query}}$ denotes the loss function for the query set, and $\phi^{\prime}$ represents the updated parameters after learning from the support set $\mathcal{T}_i^{support}$.

\subsection{The Extraction of Meta-Knowledge from Knowledge Graphs}\label{sec:meta-knowledge in KG}
Meta-knowledge refers to knowledge about knowledge, typically involving high-level abstractions of data, structures, and relationships. In the context of KGs, meta-knowledge often describes the rules, structures, and reasoning logic inherent to the graph itself, such as the types of entities and the properties of relationships. Unlike the high-level rules learned through cross-task meta-training, which are shared across tasks, the meta-knowledge within a KG can be regarded as a localized instantiation of high-level meta-knowledge. In MetaEU, we utilize two modules, the Relation-Aware Entity Embedding Generator (RAEEG) and the Neighbor-Enhanced Embedding Modulator (NEEM), to extract meta-knowledge from the KG.
\subsubsection{Relation-Aware Entity Embedding Generator}
In a KG, the outgoing relations and ingoing relations of an entity implicitly encode information about the entity's type. As shown in Figure~\ref{fig1}, even though \textit{Entity 1} is unseen, its outgoing relation \textit{works\_at} and ingoing relation \textit{father\_of} indicate that \textit{Entity 1} is likely an employee of some organization and the child of a specific father. To facilitate representation, we denote the outgoing relation embedding matrix of the KG as $\mathbf{R}^{\text{out}} \in \mathbb{R}^{|\mathcal{R}| \times d}$ and the ingoing relation embedding matrix as $\mathbf{R}^{\text{in}} \in \mathbb{R}^{|\mathcal{R}| \times d}$, where $\mathbf{R}_r^{\text{out}}$ represents the embedding of a specific relation $r$.

In the KG $\mathcal{G} = (\mathcal{E}, \mathcal{R}, \mathcal{T})$, the RAEEG module can generate the initial embedding $\mathbf{E}_e^{\text{init}}$ of an entity $e$ based on its outgoing and ingoing relations:
\begin{equation}
    \mathbf{E}_e^{init}=\frac{\sum_{r\in \mathcal{O}(e)}\mathrm{R}_{r}^{\mathrm{out}}+\sum_{r\in \mathcal{I}(e)}\mathrm{R}_{r}^{\mathrm{in}}}{|\mathcal{O}(e)|+|\mathcal{I}(e)|},
\end{equation}
where $\mathcal{I}(e) = \{r \mid \exists x, (x, r, e) \in \mathcal{T}\}$ denotes the set of ingoing relations of entity $e$, and $\mathcal{O}(e) = \{r \mid \exists x, (e, r, x) \in \mathcal{T}\}$ denotes the set of outgoing relations of entity $e$.
\subsubsection{Neighbor-Enhanced Embedding Modulator}
After RAEEG generates the initial embedding $\mathbf{E}_e^{init}$ based on the incoming and outgoing relations of entity $e$, the NEEM module captures the multi-hop neighborhood information of $e$ to further refine $\mathbf{E}_e^{init}$. Existing studies have shown that Graph Neural Networks (GNNs) are capable of capturing the local structures of knowledge graphs~\cite{schlichtkrullModelingRelationalData2018,teruInductiveRelationPrediction2020}. Therefore, the design of the NEEM module draws inspiration from the principles of GNNs. Following the structure of R-GCN~\cite{schlichtkrullModelingRelationalData2018}, the NEEM module for an entity $e$ is formulated as follows:
\begin{equation}
    \mathbf{E}_e^{(l+1)}=\sigma\left(\sum_{r\in\mathcal{R}}\sum_{e_n\in\mathcal{N}_r(e)}\frac1{c_{e,r}}W_r^{(l)}\mathbf{E}_{e_n}^{(l)}+W_0^{(l)}\mathbf{E}_e^{(l)}\right),
\end{equation}
where $\mathbf{E}_e^{(l)}$ represents the feature representation of node $e$ at layer $l$, $\mathcal{R}$ denotes the set of relations in $\mathcal{G} = (\mathcal{E}, \mathcal{R}, \mathcal{T})$, $\mathcal{N}_r(e)$ represents the set of neighboring nodes connected to $e$ via relation $r$, and $c_{e,r}$ is a normalization factor used to handle imbalanced node degrees, typically defined as $c_{e,r} = |\mathcal{N}_r(e)|$. $W_r^{(l)}$ is the weight matrix for relation $r$ at layer $l$, and $W_0^{(l)}$ is the self-loop weight matrix for processing the features of node $e$ itself. $\sigma$ represents the activation function, for which we use ReLU. The input representation for NEEM is set as $\mathbf{E}_e^{0} = \mathbf{E}_e^{\text{init}}$.

To enable the model to flexibly leverage both low-order and high-order neighborhood information and to adaptively utilize the most suitable neighborhood information for each entity, we incorporate a hierarchical embedding integrator (HEI) into NEEM. The formula is as follows:
\begin{equation}
    \mathbf{E}_e^{final}=\mathrm{HEI}(\{\mathbf{E}_e^l\}_{l=0}^L)=\mathbf{W}^\mathrm{HEI}\bigoplus_{l=0}^L\mathbf{E}_e^l,
\end{equation}
where $\mathbf{E}_e^{final}$ represents the final embedding of entity $e$; $\mathrm{HEI}(\{\mathbf{E}_e^l\}_{l=0}^L)$ denotes the HEI module, which is responsible for hierarchically integrating all embeddings from layer $0$ to layer $L$; $\bigoplus$ represents the layer-wise concatenation operation; and $\mathbf{W}^\mathrm{HEI}$ refers to the transformation matrix in HEI, which maps the concatenated high-dimensional representation into the final entity embedding.
\subsection{Ensemble Learning and Ensemble Unlearning}
This section introduces the ensemble learning and ensemble forgetting mechanisms in the MetaEU framework. As shown in Figure~\ref{fig2}, by integrating multiple base models, the framework decomposes the overall learning objective into smaller, independently solvable tasks handled by each base model. The base models combine the meta-knowledge learned from cross-task meta-training in Section~\ref{sec:framework of meta-learning} with the meta-knowledge extracted from the knowledge graph in Section~\ref{sec:meta-knowledge in KG}, enabling the generation of high-quality embeddings and high-quality obfuscated data. This allows the model to perform well on the retention set while minimizing the influence of data from the forgetting set. In the knowledge graph $\mathcal{G} = (\mathcal{E}, \mathcal{R}, \mathcal{T})$, the objective function for ensemble learning can be expressed as:
\begin{equation}
    \arg\min_{f^{base},w}\mathcal{L}_{ensembleL}=\sum_{i=1}^Nw_i\mathcal{L}(f_i^{base},\mathcal{T}_{query}),
\end{equation}
where $f^{\text{base}}$ represents a base model, $w_i$ denotes the weight assigned to the $i$-th base model, satisfying $\sum_{i=1}^N w_i = 1$; $N$ represents the number of base models; and $\mathcal{L}_{ensembleL}$ refers to the ensemble learning loss function (the calculation method can be referenced from Equation~\ref{KGE loss}). As shown in Figure~\ref{fig2}, $L_1$ is an example of such a loss. The objective of $L_1$ is to increase the KGE scores of positive samples in the query set $\mathcal{T}_{query}$ and decrease the KGE scores of negative samples through backward.

Correspondingly, the objective function for ensemble unlearning can be expressed as:
\begin{equation}
    \arg\max_{f^{base},w}\mathcal{L}_{ensembleU}=\sum_{i=1}^Nw_i\mathcal{L}(f_i^{base},\mathcal{T}_{query}),
    \label{ensemble unlearning}
\end{equation}
where $\mathcal{L}_{ensembleU}$ represents the loss function for ensemble unlearning, corresponding to $L_2$ in Figure~\ref{fig2}. The objective of Equation~\ref{ensemble unlearning} is to maximize the overall loss in the query set of the forgetting set by adjusting the base models $f^{base}$ and their weights $w$, thereby achieving the effect of forgetting specific knowledge.

$L_3$ in Figure~\ref{fig2} guides the optimization of embeddings by combining $L_1$ and $L_2$ with weights; its formula is:
\begin{equation}
    L_3=w_aL_1+w_bL_2,\quad\mathrm{s.t.} w_a+w_b=1, w_a,w_b\in[0,1].
\end{equation}
$L_4$ is used to balance the model's \textit{forgetting strength} during the ensemble unlearning process, in order to prevent excessive unlearning from affecting the overall performance of the model. The role of $L_5$ is to fine-tune the model for specific knowledge graphs in particular tasks.
\section{Experiment}
\subsection{Information about the Dataset}
Following existing work~\cite{zhuHeterogeneousFederatedKnowledge2023}, we use the FB15k-237 dataset in this experiment to evaluate the effectiveness of MetaEU. According to the description of the meta-learning framework in Section~\ref{sec:framework of meta-learning}, we need to extract $k$ subgraphs $\mathcal{G}_{i} = (\mathcal{E}_{i}, \mathcal{R}_{i}, \mathcal{T}_{i})$ from the graph $\mathcal{G}$ to correspond to $k$ tasks $T_i$. Furthermore, each subgraph triplet needs to be divided into a support set $\mathcal{T}_i^{\text{support}}$ and a query set $\mathcal{T}_i^{\text{query}}$, and the relationship between the two parts is as shown in Equation~\ref{eq:support set query set}.
\subsection{Experimental Setup}
\subsubsection{Implementation Details}
We implement MetaEU using PyTorch and DGL, and the experiments were conducted on an Intel(R) Xeon(R) Gold 6226R CPU and an NVIDIA GeForce RTX 3090 GPU. The learning rate is set to $10^{-2}$. The number $k$ of tasks $T_i$ and their corresponding subgraphs $\mathcal{G}_i$ is set to 10,200, where the number of training tasks is 10,000 and the number of validation tasks is 200. The number of epochs for meta-training is set to 10, and the batch size is set to 64. For the NEEM module, the number of layers $l$ is set to 3.
\subsubsection{Evaluation Metrics}
We independently conduct the experiments 10 times and compute the average to obtain the final reported results. We use Hits@n and MRR as evaluation metrics to validate the performance of the MetaEU model. Hits@n measures the average proportion of knowledge ranked within the top n in link prediction, while MRR calculates the mean reciprocal rank of the predictions. For the evaluation metrics used in this experiment, higher numerical values indicate better model performance on the corresponding tasks (e.g., link prediction). The calculation methods for the aforementioned metrics are defined as follows:
\begin{equation}
    \mathrm{Hit}@n=\frac{1}{|Q|}\sum_{q\in Q}\mathbb{I}(r_{q}\leq n),
\end{equation}
\begin{equation}
    \mathrm{MRR}=\frac1{|Q|}\sum_{q\in Q}\frac1{r_q},
\end{equation}
where $|Q|$ is the total number of queries or test instances. $r_{q}$ is the rank position of the relevant item for query $q$. $\mathbb{I}(\cdot)$ is the indicator function, which equals 1 if the condition inside is true and 0 otherwise.
\subsection{Baseline}
\begin{figure}
  % \vspace{-0.5cm} % 减少图形和文字之间的间隔
  \centering
  \includegraphics[width=0.8\textwidth]{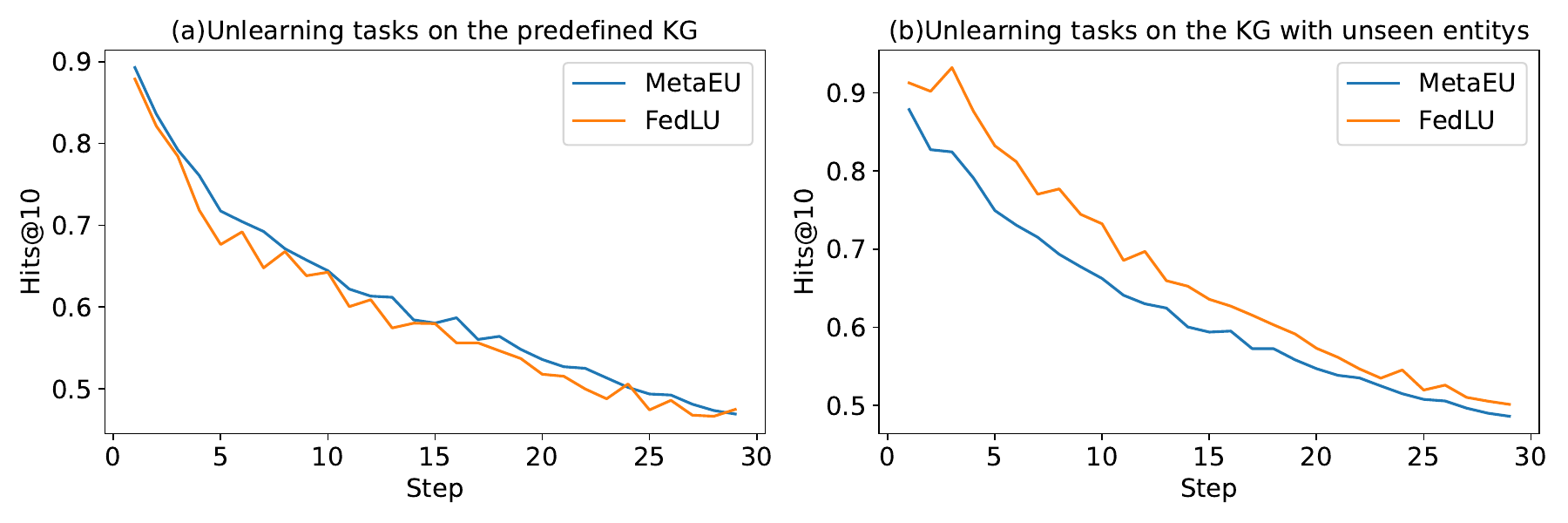}
  \caption{Comparison of the performance of MetaEU and FedLU on the unlearning task.}
  \label{fig:metaeu-fedlu}
  % \vspace{-0.5cm} % 减少图形和文字之间的间隔
\end{figure}

Existing KGE unlearning works are relatively few and differ significantly in task context from our proposed method. The existing approaches~\cite{zhuHeterogeneousFederatedKnowledge2023,liuFederatedKnowledgeGraph2024} rely on federated learning frameworks and perform KGE unlearning tasks on a fixed, predefined knowledge graph (where all entities in the KG are previously seen). In contrast, MetaEU employs multi-task training and learns meta-knowledge based on relation types and entity categories within the KG, enabling it to perform unlearning tasks on knowledge graphs containing unseen entities. As shown in Figure~\ref{fig:metaeu-fedlu}, although MetaEU's performance in Figure~\ref{fig:metaeu-fedlu}(a) is not outstanding, in Figure~\ref{fig:metaeu-fedlu}(b), the decline in MetaEU's Hits@10 metric is much faster than that of FedLU~\cite{zhuHeterogeneousFederatedKnowledge2023}. This may be because traditional KGE unlearning models cannot generate suitable embeddings for unseen entities in the early stages of training, whereas MetaEU addresses this issue more effectively.
\subsection{Main Results and Analysis}
As shown in Table~\ref{tab:metaeu four kge}, to demonstrate the effectiveness and generalizability of the MetaEU framework, we conducted experiments on four representative KGE models~\cite{bordesTranslatingEmbeddingsModeling2013,yangEmbeddingEntitiesRelations2014,trouillonComplexEmbeddingsSimple2016,sunRotatEKnowledgeGraph2018}. "RAW" represents the embeddings obtained by training the model on the original complete dataset before removing the forgetting set; "Retrained" represents the embeddings obtained by training the model on the remaining dataset after removing the forgetting set; "Unlearned" represents the embeddings obtained by applying unlearning to the RAW embeddings. Based on the above results, we found that: (i) RAW achieved the highest scores on the Hits@n and MRR metrics for both the Test and Forget sets; (ii) Retrained, after removing the forgetting set, had lower scores on the Test set compared to RAW, but higher scores on the Forget set than Unlearned. This suggests that simply removing the forgetting set and retraining the model does not fully mitigate the influence of the forgetting set; (iii) Unlearned exhibited the lowest performance on the Forget set but showed performance on the Test set that was comparable to RAW. This indicates that, through unlearning, the model effectively reduces the impact of the forgetting set while preserving its performance on the remaining set.
\begin{table}[ht]
    \centering
    \caption{MetaEU works effectively across four representative KGE models, demonstrating the framework's efficacy and versatility.}
    \begin{tabular}{lcccccccc}
        \toprule
        & \multicolumn{2}{c}{TransE} & \multicolumn{2}{c}{DistMult} & \multicolumn{2}{c}{ComplEx} & \multicolumn{2}{c}{RotatE} \\
        \cmidrule(lr){2-3} \cmidrule(lr){4-5} \cmidrule(lr){6-7} \cmidrule(lr){8-9}
        & Test $\uparrow$ & Forget $\downarrow$ & Test $\uparrow$ & Forget $\downarrow$ & Test $\uparrow$ & Forget $\downarrow$ & Test $\uparrow$ & Forget $\downarrow$ \\
        \midrule
        \multicolumn{7}{l}{\textbf{RAW}} \\
        MRR & 0.7254 & 0.7157 & 0.7065 & 0.7104 & 0.7141 & 0.7057 & 0.7241 & 0.7226 \\
        Hits@1 & 0.6055 & 0.6035 & 0.5793 & 0.5650 & 0.5889 & 0.5935 & 0.6017 & 0.6011 \\
        Hits@5 & 0.8815 & 0.8713 & 0.8744 & 0.8648 & 0.8796 & 0.8713 & 0.8853 & 0.8901 \\
        Hits@10 & 0.9586 & 0.9422 & 0.9596 & 0.9601 & 0.9907 & 0.9892 & 0.9651 & 0.9652 \\
        \midrule
        \multicolumn{7}{l}{\textbf{Retrained}} \\
        MRR & 0.6918 & 0.2473 & 0.6791 & 0.2269 & 0.6825 & 0.2361 & 0.7012 & 0.2350 \\
        Hits@1 & 0.5831 & 0.1878 & 0.5527 & 0.1563 & 0.5563 & 0.1396 & 0.5774 & 0.2037 \\
        Hits@5 & 0.8593 & 0.3475 & 0.8396 & 0.2988 & 0.8458 & 0.3087 & 0.8499 & 0.4328 \\
        Hits@10 & 0.9375 & 0.4734 & 0.9357 & 0.4271 & 0.9582 & 0.4265 & 0.9422 & 0.5912 \\
        \midrule
        \multicolumn{7}{l}{\textbf{Unlearned}} \\
        MRR & 0.7153 & 0.1740 & 0.6902 & 0.1853 & 0.6910 & 0.1812 & 0.7124 & 0.1983 \\
        Hits@1 & 0.5941 & 0.1041 & 0.5620 & 0.0947 & 0.5721 & 0.0984 & 0.5879 & 0.1028 \\
        Hits@5 & 0.8673 & 0.2203 & 0.8580 & 0.2534 & 0.8543 & 0.2748 & 0.8741 & 0.2587 \\
        Hits@10 & 0.9426 & 0.3607 & 0.9531 & 0.3877 & 0.9626 & 0.3930 & 0.9584 & 0.3792 \\
        \bottomrule
    \end{tabular}
    \label{tab:metaeu four kge}
\end{table}
\subsection{Ablation Study}
\begin{figure}
  % \vspace{-0.5cm} % 减少图形和文字之间的间隔
  \centering
  \includegraphics[width=0.8\textwidth]{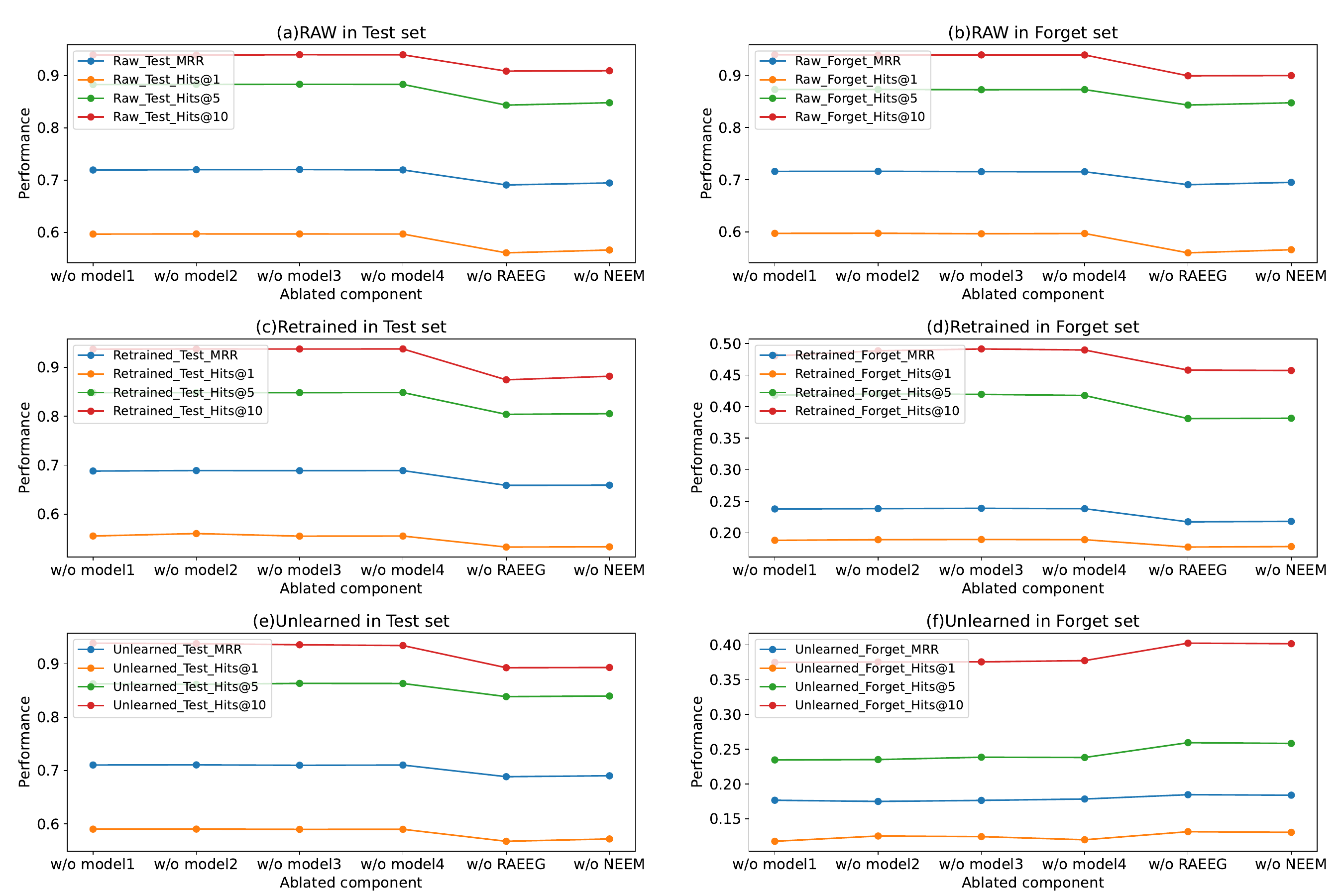}
  \caption{Ablation studies on different components of MetaEU.}
  \label{fig:ablation}
  % \vspace{-0.5cm} % 减少图形和文字之间的间隔
\end{figure}

We conduct ablation studies on different components of MetaEU. For the ensemble learning and ensemble unlearning components, we sequentially ablate models 1 to 4. To ablate the RAEEG component, we randomly initialize the entity embeddings during the training process. To ablate the NEEM component, we omit the step of enhancing entity embeddings through neighboring nodes and directly use the embeddings generated by RAEEG as the entity embeddings. As shown in Figure~\ref{fig:ablation}, we observe that when any base model is ablated within the ensemble learning or ensemble unlearning components, the model's performance does not show significant changes. This indicates that the ensemble model is robust, and there is some redundancy within the base learners and base unlearners. Additionally, we find that after removing the RAEEG or NEEM components, there is a noticeable decline in performance on the Test set for all embeddings. However, the Unlearned embeddings show an improvement on the Forget set, which suggests that RAEEG and NEEM can help the model perform better on the KGE unlearning task.
\section{Conclusion}
We propose MetaEU, a novel knowledge graph embedding unlearning model based on meta-learning framework. This framework leverages meta-training regime and the entity types and relational properties within the knowledge graph to learn meta-knowledge that is independent of specific unlearning task scenarios. By incorporating ensemble learning and ensemble unlearning processes, MetaEU can eliminate the influence of specific knowledge on the KGE model while preserving the overall performance of the model. Our experimental results demonstrate that MetaEU can efficiently perform KGE unlearning in unfamiliar scenarios with unseen entities, a capability that existing methods lack. In future work, we will explore the application of meta-learning-based KGE unlearning methods in multi-source knowledge graph fusion and investigate more effective KGE unlearning approaches.
\subsubsection{\ackname} This study was funded by X (grant number Y).

\bibliographystyle{splncs04_unsort} 
\bibliography{ref1124}

% -----下面的内容是手动创建参考文献列表，是最简单的方法，适用于文献不太多的情况




% LNCS Homepage, \url{http://www.springer.com/lncs}, last accessed 2023/10/25
% \end{thebibliography}

\end{document}